\documentclass[11pt]{article}
\usepackage[final]{acl}   
\hypersetup{
  pdftitle={Enhancing Legal Reasoning in Pre-trained Language Models via Knowledge Graph-Guided Multi-Task Pre-training},
  pdfauthor={Ibrahim Siddig, Munir Georges}
}

\usepackage{inconsolata}

\usepackage{times}
\usepackage{latexsym}
\usepackage[T1]{fontenc}
\usepackage[utf8]{inputenc}
\usepackage{microtype}
\usepackage{graphicx}

\usepackage{amsmath}
\usepackage{amssymb}

\usepackage{booktabs}
\usepackage{multirow}
\usepackage{array}
\usepackage{threeparttable}

\usepackage{enumitem}
\usepackage{subcaption}
\usepackage{float}

\usepackage{pgfplots}
\usepackage{tikz}
\usetikzlibrary{positioning, arrows.meta, calc}
\pgfplotsset{compat=1.18}

\usepackage{tcolorbox}
\newtcolorbox[auto counter, number within=section]{examplebox}[2][]{
  colback=blue!4!white,
  colframe=blue!60!black,
  title={Box~\thetcbcounter: #2},
  fonttitle=\bfseries\small,
  sharp corners, boxrule=0.6pt,
  left=4pt, right=4pt, top=3pt, bottom=3pt,
  #1
}

\title{Enhancing Legal Reasoning in Pre-trained Language Models via\\
       Knowledge Graph-Guided Multi-Task Pre-training}

\author{
  Ibrahim Siddig \and Munir Georges \\
  AImotion Bavaria, Technische Hochschule Ingolstadt \\
  \texttt{\{ibrahim.siddig, munir.georges\}@thi.de} \\
}

\begin{document}
\maketitle

\begin{abstract}

Legal language understanding requires reasoning over semi-structured regulatory documents, yet standard pre-training objectives do not explicitly target this structural capability. 
Because current legal language models rely solely on masked language modeling,
they treat statutory text as flat token sequences and ignore the underlying hierarchical constraints. 
To bridge this gap, we present a multi-task pre-training approach for German traffic law (\textit{StVO}). 
Our approach combines three complementary objectives: Next Sentence Prediction (NSP), 
Knowledge Graph~(\(KG\)) Guided Clustering (CLU), and Knowledge Graph~(\(KG\)) Aware Paragraph Classification (CLS).
This approach explicitly encodes structured legal knowledge directly into the model weights. 
This encoding enables zero-shot legal reasoning at inference time and removes the need for discrete graph traversal. 
We evaluate our approach across nine Transformer-based encoders and 324 experimental runs. 
On synthetic \textit{StVO} question answering, our models demonstrate consistent relative improvements between 10.3\% and 155.0\% over baselines. 
Furthermore, relative to the established LexGLUE baselines~\citep{chalkidis-etal-2022-lexglue}, our models improve over prior results on four of seven legal benchmarks.
It also achieves consistent zero-shot transfer gains between 3.4\% and 15.8\% on Austrian driving regulations. These experimental results indicate that our training approach generalizes successfully beyond the initial training legal domain.

\end{abstract}

\section{Introduction}
\label{sec:intro}

Legal Natural Language Processing~(NLP) models are considered a base component 
for Reliable Artificial Intelligence~(rAI) within semi-structured resources 
such as statutory books and contractual texts~\citep{yao2024lawyer}. 
A persistent challenge involves the trade-off between predictive accuracy and 
interpretability; as SOTA Pre-trained Language Models~(PLMs) achieve high performance, they remain largely unexplainable; conversely, rule-based methods offer transparency but suffer from flexibility~\citep{zhong-etal-2020-nlp}. Despite recent proposals targeting this balance~\citep{An2024ResearchOI,Wang2024LegalReasonerAM}, 
the gap between empirical performance and verifiable legal reasoning persists 
as a critical bottleneck~\citep{zhong-etal-2020-nlp}.

Existing legal domain-adaptive pre-training approaches---including 
Lawformer~\citep{xiao2021lawformer}, SAILER~\citep{li2023sailer}, and 
Legal-BERT~\citep{chalkidis2020legalbert}---rely on implicit structural 
modeling through long-range attention, structural headers, or domain-specific 
MLM, respectively. None \emph{explicitly} incorporates the formal semantic 
relationships and hierarchical constraints of the semi-structured legal texts 
during pre-training. To resolve this limitation, we introduce a training approach that explicitly injects knowledge graphs into the embedding space of PLMs.

In this work, we propose a multi-task domain-adaptive pre-training 
approach validated in the German Road Traffic Regulations~(\textit{StVO}) as representative for the semi-structured legal texts, encoding legal structure via three complementary objectives: Next Sentence Prediction~(NSP) for document coherence, Knowledge Graph-guided Clustering~(CLU) for relational structure, and \(KG\)-aware paragraph Classification~(CLS) for task-aligned reasoning.

Our main contributions in this work are:
\begin{itemize}[leftmargin=*, itemsep=0.15em, topsep=2pt]
    \item We introduce a unified multi-task pre-training approach for semi-structured regulatory text that explicitly integrates KG structures into latent representations, unlike existing text-concatenation approaches.
    
    \item We employ TransE-derived legal paragraph embeddings as \emph{soft} pre-training targets, enabling the transfer of relational legal knowledge without hard classification constraints and allowing the model to learn a continuous 
    range of legal concepts. 

    \item A differentiable gated \(KG\)-aware attention training mechanism that performs real-time latent-space retrieval from a \(KG\) paragraph embeddings library, dynamically weighting external legal context against internal model 
    representations. Unlike traditional retrieval-augmented approaches, which require discrete external database lookups during inference.

    \item A synthetic German traffic law QA dataset ($N=12{,}554$) with 
    paragraph-level citation grounding and strict paragraph-level splits to 
    prevent data leakage, providing a new benchmark for citation-accurate 
    legal reasoning.
\end{itemize}

\section{Related Work}
\label{sec:related}

\paragraph{Legal Language Models.}
Domain-specific Transformer adaptations have shaped legal NLP. 
\citet{chalkidis2020legalbert} established \textit{Legal-BERT} via continued 
pre-training on English legal corpora. \citet{xiao2021lawformer} introduced 
\textit{Lawformer} for long-range dependencies. \citet{li2023sailer} proposed 
\textit{SAILER} for structure-aware dense retrieval. These models rely on raw 
linguistic sequences and do not explicitly incorporate structured relational knowledge during pre-training. They remain structurally blind to underlying semantic dependencies. On the evaluation side, LexGLUE~\citep{chalkidis-etal-2022-lexglue} standardized assessment across seven English legal tasks, extended by \citet{chalkidis-etal-2022-fairlex} for fairness and \citet{t-y-s-s-etal-2024-lexsumm} for summarization. LegalBench~\citep{Guha2023LegalBenchAC} provides legal English QA samples but lacks German regulatory coverage and paragraph citation sourcing.

\paragraph{Knowledge Graph-Enhanced Embeddings.}
The transition toward neuro-symbolic integration has driven structure-aware 
pre-training in legal NLP~\citep{zhang2019ernie,li2023sailer}. 
\citet{yao2019kg} proposed \textit{\(KG\)-BERT}, treating triples as text 
sequences to leverage linguistic patterns for graph completion. 
\citet{xie2016representation} and \citet{alam2025towards} argue that standard 
embeddings overlook schematic information and logical axioms, suggesting 
ontology-enriched representations for deep semantic understanding. 
\citet{tang2024graphgpt} introduced \textit{GraphGPT}, aligning structural 
graph patterns with natural language via dual-stage instruction tuning and 
Chain-of-Thought distillation for zero-shot relational reasoning. We build our approach on \textbf{TransE}~\citep{bordes2013translating,wang2014knowledge} specifically for its efficiency in preserving hierarchical topology as soft targets for the CLU objective. In preliminary experiments, we evaluated a diverse range of \(KG\) embeddings—including bilinear models (\textbf{DistMult},~\citep{yang2015embedding}), 
complex-valued embeddings (\textbf{ComplEx},~\citep{pmlr-v48-trouillon16}),
and rotational models (RotatE, \citealp{sun2018rotate})---with TransE
yielding the most robust downstream performance (Table~\ref{tab:kge-ablation}). This approach bypasses the computational costs typically associated with Transformer-based graph encoders \citep{yao2019kg}.

\begin{table}[t]
\centering
\small
\begin{tabular}{lc}
\toprule
\textbf{KG embedding} & \textbf{m-F1} \\
\midrule
TransE   & \textbf{0.864} \\
RotatE   & 0.532 \\
ComplEx  & 0.392 \\
DistMult & 0.370 \\
\bottomrule
\end{tabular}
\caption{Downstream StVO QA performance (m-F1) of the KG-guided model on
\texttt{bert-base-echr}, with the pre-training configuration fixed and only the
KG embedding varied. TransE yields the most robust downstream performance,
motivating its use for the CLU soft targets.}
\label{tab:kge-ablation}
\end{table}

\paragraph{Multi-Task and Knowledge Distillation.}
Typical \textbf{knowledge distillation} \citep{hinton2015distilling} focuses on model compression; however, we adapt this paradigm to distill structured \(KG\) graph topology into \emph{soft pre-training targets}. Furthermore, we leverage \textbf{Multi-Task Learning} (MTL) frameworks \citep{caruana1997multitask, kendall2018multi} to treat text-based modeling and graph-based understanding as synergistic objectives. Consequently, our work extends beyond traditional text augmentation by proposing an architecture in which model embeddings are intrinsically aligned with the structured logical hierarchies of a legal knowledge graph.

\section{Methodology}
\label{sec:methodology}

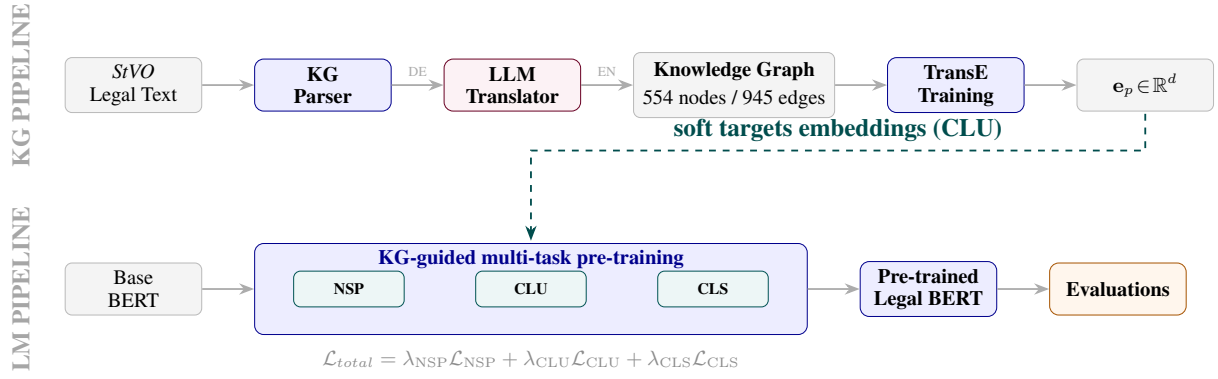
\begin{figure*}[t]
\centering

\resizebox{\textwidth}{!}{%
\begin{tikzpicture}[
    font=\small,
    node distance=0.5cm and 0.8cm, 
    mybox/.style={rectangle, rounded corners=3pt,
        draw=gray!55, fill=gray!9,
        minimum width=2.1cm, minimum height=0.8cm,
        align=center, inner sep=3pt},
    graphbox/.style={rectangle, rounded corners=3pt,
        draw=gray!55, fill=gray!9,
        minimum width=2.3cm, minimum height=1.0cm,
        align=center, inner sep=4pt},
    mystage/.style={rectangle, rounded corners=4pt,
        draw=blue!55!black, fill=blue!7,
        minimum width=2.1cm, minimum height=0.8cm,
        align=center, inner sep=3pt, font=\small\bfseries},
    mytask/.style={rectangle, rounded corners=3pt,
        draw=teal!65!black, fill=teal!7,
        minimum width=1.7cm, minimum height=0.55cm,
        align=center, inner sep=2pt, font=\scriptsize\bfseries},
    myresult/.style={rectangle, rounded corners=4pt,
        draw=orange!65!black, fill=orange!7,
        minimum width=2.1cm, minimum height=0.8cm,
        align=center, inner sep=3pt, font=\small\bfseries},
    myarr/.style={-Stealth, thick, gray!65},
    mydarr/.style={-Stealth, thick, teal!60!black,
                    dashed, shorten >=1pt, shorten <=1pt},
    mylbl/.style={font=\footnotesize, text=gray!70, font=\bfseries},
]

\node[mybox]   (stvo)   {\textit{StVO}\\Legal Text};
\node[mystage, right=of stvo]   (kgb)    {KG\\Parser};
\node[mystage, right=of kgb, draw=purple!55!black, fill=purple!5] (trans) {LLM\\Translator};
\node[graphbox, right=of trans]  (graph)  {\textbf{Knowledge Graph}\\[2pt]\small 554 nodes / 945 edges};
\node[mystage, right=of graph]  (transe) {TransE\\Training};
\node[mybox,   right=of transe] (emb)    {$\mathbf{e}_p\!\in\!\mathbb{R}^{d}$};

\draw[myarr] (stvo)--(kgb); 
\draw[myarr] (kgb)--(trans) node[midway, above, font=\tiny] {DE};
\draw[myarr] (trans)--(graph) node[midway, above, font=\tiny] {EN};
\draw[myarr] (graph)--(transe); 
\draw[myarr] (transe)--(emb);
\node[mylbl, left=0.4cm of stvo, rotate=90, anchor=south]{KG PIPELINE};

\node[mybox,   below=2.3cm of stvo] (bert) {Base\\BERT};
\node[mystage, right=of bert, minimum width=8.5cm, minimum height=1.4cm] (mtp) {};
\node[font=\footnotesize\bfseries, text=blue!55!black] at (mtp.north) [yshift=-0.25cm] {KG-guided multi-task pre-training};

\node[mytask] (nsp) at ([xshift=-2.8cm]mtp.center) {NSP};
\node[mytask] (clu) at ([xshift= 0.0cm]mtp.center) {CLU};
\node[mytask] (cls) at ([xshift= 2.8cm]mtp.center) {CLS};

\node[mystage,  right=of mtp]    (ptbert) {Pre-trained\\Legal BERT};
\node[myresult, right=of ptbert] (eval)    {Evaluations};

\draw[myarr] (bert)--(mtp); 
\draw[myarr] (mtp)--(ptbert);
\draw[myarr] (ptbert)--(eval);
\node[mylbl, left=0.4cm of bert, rotate=90, anchor=south]{LM PIPELINE};

\draw[mydarr] (emb.south) -- ++(0,-0.6) -| (mtp.north -| clu.north)
    node[pos=0.25, above, font=\footnotesize, text=teal!60!black, font=\bfseries]{soft targets embeddings (CLU)};

\node[font=\footnotesize, text=gray!90, anchor=north] at (mtp.south) [yshift=-0.1cm] 
    {$\mathcal{L}_{total} = \lambda_{\mathrm{NSP}}\mathcal{L}_{\mathrm{NSP}} + \lambda_{\mathrm{CLU}}\mathcal{L}_{\mathrm{CLU}} + \lambda_{\mathrm{CLS}}\mathcal{L}_{\mathrm{CLS}}$};
\end{tikzpicture}%
}

\vspace{-4pt}
\caption{Architecture of the proposed KG-guided multi-task pre-training pipeline. German \textit{StVO} legal text is parsed, translated via an LLM translator, and structurally aligned with a corresponding English-based knowledge graph to inject explicit relational constraints into model reasoning. A detailed visualization of the generated graph topology is provided in Appendix~\ref{sec:appendix_kg_viz}.}
\label{fig:pipeline}
\end{figure*}

Our methodology, depicted in Figure~\ref{fig:pipeline}, follows a hierarchical progression across two converging pipelines.

\begin{itemize}[leftmargin=*, itemsep=0.3em]
    \item \textbf{The \(\textbf{KG}\) Pipeline:} This workflow transforms raw \textit{StVO} legal text into a formal \textbf{Knowledge Graph} via the legal pattern-based parser \citep{Siddig2025_1260}. These relationships are mapped into a latent translation space using \textbf{TransE}, generating paragraph embeddings ($\mathbf{e}_p$) that capture the topological core of the legal code.
    
    \item \textbf{The LM Pipeline:} This workflow begins with a \textbf{pre-trained base encoder}. It then enters a \textbf{domain-adaptive} stage, where the model undergoes \textbf{\(\textbf{KG}\)-guided multi-task pre-training}. In this stage, the model internalizes structured legal logic before proceeding to final \textbf{Evaluations}.
\end{itemize}

By synthesizing these two pipelines, the architecture moves beyond simple surface-level text patterns, ensuring that the underlying regulatory structure of German traffic law (\textit{StVO}) is fundamentally injected into the model's representations.

\subsection{Problem Formulation}
\label{subsec:problem_formulation}
Given \textit{StVO} legal document $\mathcal{D}$, we train representations that enable accurate legal reasoning on question-answering. Formally, for a legal question and a possible answer pair represented in a sequence of words $(q, a)$ and its source  legal paragraph $p$, we learn a function $f: (q,a,p) \rightarrow \{0, 1\}$ indicating whether the answer is legally correct given the paragraph. The training approach incorporates three complementary objectives in Section \ref{subsec:multitask_pretraining}:
(1)~NSP for document coherence,
(2)~CLU for KG-guided semantic clustering, and
(3)~CLS for task-aligned binary classification.

\subsection{TransE Paragraph Embeddings Generation}
\label{subsec:transE_kg_embeddings}

We parse \textit{StVO} into a knowledge graph
$\mathcal{G}=(\mathcal{V},\mathcal{E},\mathcal{R})$ via the
parser~\citep{Siddig2025_1260} exploiting the
semi-structured hierarchy of the German traffic law (categories,
paragraphs, sub-paragraphs, sentences, tables, sections, signs). The graph contains \textbf{554 nodes} and \textbf{945 edges} across five different relations: \textsc{has\_paragraph}, \textsc{has\_sentence}, \textsc{has\_section}, \textsc{has\_sign}, \textsc{has\_table}. TransE learns a single embedding per entity node, independent of whether the node appears as the head or tail of a triple. We use the resulting paragraph-node embeddings $\mathbf{e}_p\in\mathbb{R}^{500}$ as \emph{soft clustering targets} for the CLU objective (Section~\ref{subsec:multitask_pretraining}). They preserve relational structure and enable gradient flow through cosine similarity.

\subsection{Hierarchical Document Encoding}

To handle the structural complexity and length of the legal sentences in the \textit{StVO}, we adopt the idea of a two-level hierarchical architecture \citep{yang2016hierarchical, chalkidis2019neural}. Each legal sentence is decomposed into an ordered set of segments $\{s_1, \dots, s_K\}$ by a heuristic structure-aware segmenter. The segmenter applies a cascade of manually designed patterns that split text
at the statutory structural markers of the StVO: at paragraph markers (\S), then at sub-paragraph markers ((1), (2), \dots), and finally, at sentence boundaries for legal sentences exceeding the encoder's token limit. First, a shared \textbf{segment encoder} $\mathcal{E}_{\text{base}}$ processes individual segments $s_k$ to produce local embeddings: $\mathbf{h}_k = \mathcal{E}_{\text{base}}(s_k)$. Second, a \textbf{Document Encoder} $\mathcal{E}_{\text{doc}}$ aggregates the sequence $\mathbf{H} = [\mathbf{h}_1, \dots, \mathbf{h}_K]$ to capture global dependencies, yielding the final document representation $\mathbf{z}_d = \text{MaxPool}(\mathcal{E}_{\text{doc}}(\mathbf{H}))$. 
This design preserves intra-document structural context while remaining computationally efficient for extensive statutory sequences.

\subsection{Multi-Task Pre-Training Objectives}
\label{subsec:multitask_pretraining}

\paragraph{Task 1 --- Next Sentence Prediction~(NSP)}
we employ domain-adapted Next Sentence Prediction (NSP) \citep{devlin2019bert}. Positive pairs are drawn from consecutive sentences within the same legal paragraph, while negative pairs are sampled from random cross-paragraph sequences. Given a pair of legal sentences $(s_i, s_j)$ and a binary label $y \in \{0, 1\}$ indicating whether $s_j$ is the actual successor of $s_i$, the objective is to minimize the binary cross-entropy loss:
\begin{equation}
\begin{split}
    \mathcal{L}_{\mathrm{NSP}} = - \sum [ &y \log(P(y=1|s_i, s_j)) + \\
    &(1-y) \log(1 - P(y=1|s_i, s_j)) ]
\end{split}
\end{equation}
where $P(y=1|s_i, s_j)$ represents the probability predicted by the model that the sentences are consecutive.

\paragraph{Task 2 --- \(KG\)-Guided Clustering~(CLU).}
For each legal segment $s$, the hierarchical encoding produces a representation $\mathbf{z}_s \in \mathbb{R}^d$.
TransE pre-training has already positioned every paragraph $p$
at a learned coordinate $\mathbf{e}_p \in \mathbb{R}^{500}$
in a continuous space where structurally related paragraphs
(e.g., §\,8 and §\,9 on right-of-way) cluster together.
The CLU objective exploits this geometry as a supervision
signal via margin-based contrastive
learning~\citep{schroff2015facenet,wang2014knowledge}:
\begin{align}
  \mathcal{L}_{\text{CLU}}
    &= \underbrace{1 - \cos(\mathbf{z}_s,\,\mathbf{e}_{c^+})
       }_{\text{pull toward source paragraph}} \nonumber\\
    &\quad+\;
       \underbrace{\max\!\left(0,\;
         \cos(\mathbf{z}_s,\,\mathbf{e}_{c^-}) + m
       \right)}_{\text{push from unrelated paragraph}}
\end{align}
where $c^-\!\sim\!\text{Uniform}(\mathcal{P}\!\setminus\!\{c^+\})$
samples a random negative from the 59 \textit{StVO} paragraphs.
The \emph{positive term} minimises the cosine distance between
$\mathbf{z}_s$ and the \(KG\) embedding of its true source paragraph
$c^+$, anchoring the encoder to the correct region.
The \emph{negative term} is a hinge loss that fires only when an
unrelated paragraph $c^-$ lies within margin $m$ of $\mathbf{z}_s$,
penalising insufficient separation while ignoring already
well-separated negatives.
Crucially, because $\mathbf{e}_{c^+}$ is a 500-dimensional
continuous vector encoding structural relationships of the \(KG\),
the gradient carries relational geometry rather than discrete
isolated targets.

\paragraph{Task 3 --- \(KG\)-aware Paragraph Classification~(CLS)}
Standard classification models rely solely on the self-contained encoder representation $\mathbf{z}_s$. We relax this constraint by introducing a \textbf{dynamic paragraph attention} training mechanism that consults the \textit{StVO} paragraphs at decision time. Using $\mathbf{z}_s$ as a query, the model computes an attention distribution $\alpha$ over the full set of KG paragraph embeddings $\{\mathbf{e}_j\}$, yielding a \(KG\)-context vector $\mathbf{c}_{\text{KG}}$:
\begin{equation}
  \alpha_j = \text{softmax}\left(\frac{\mathbf{z}_s^\top \mathbf{e}_j}{\sqrt{d_h}}\right), \quad \mathbf{c}_{\text{KG}} = \sum \alpha_j \mathbf{e}_j
\end{equation}
To integrate this knowledge, a \textbf{gated fusion} step computes $\tilde{\mathbf{z}}_s = g \odot \mathbf{c}_{\text{KG}} + (1 - g) \odot \mathbf{z}_s$, where the gate $g = \sigma([\mathbf{z}_s; \mathbf{c}_{\text{KG}}]\mathbf{W}_g)$ adaptively regulates the influence of the \(KG\) paragraph embeddings by weighing the semantic compatibility between the input text and the retrieved \(KG\) context. Unlike prior work that integrates \(KG\) knowledge as static
pre-training constraints~\citep{peters2019knowledge} or via
token-level entity linking~\citep{zhang2019ernie}, our
mechanism operates at the \emph{paragraph level,} routes KG knowledge dynamically per input instance, and gates its influence to zero when the retrieved context is uninformative. This transfers the classification decisions directly to the formal \textit{StVO} during training without requiring graph traversal at inference. The final loss is described as follows:
\begin{equation}
\begin{split}
    \hat{y} &= \sigma(\mathbf{W}_c\,\tilde{\mathbf{z}}_s + b_c) \\[-0.5ex]
    \mathcal{L}_{\text{CLS}} &= -y\log\hat{y} - (1-y)\log(1-\hat{y})
\end{split}
\end{equation}
The individual objectives target different aspects of the model's capabilities. However, we unify them through a multi-task learning approach. The total objective is defined as:
\begin{equation}
\begin{split}
  \mathcal{L}_{total} = \lambda_{\text{NSP}}\mathcal{L}_{\text{NSP}} + \lambda_{\text{CLU}}\mathcal{L}_{\text{CLU}} &+ \lambda_{\text{CLS}}\mathcal{L}_{\text{CLS}} \\
  &\text{where } \textstyle\sum \lambda = 1.0
\end{split}
\end{equation}

\section{Experimental Settings}
\label{sec:setup}
To evaluate the effectiveness of our approach, we design our experiments around three main questions.
\begin{enumerate}[label=\textbf{RQ\arabic*:},
                  leftmargin=*, itemsep=0.2em, topsep=2pt]
  
  \item 
        To what extent does the semi-structured nature of
        \textit{StVO} facilitate the transfer of question-answering legal reasoning across languages, and how do domain-specific linguistic
        features impact this transfer?
  
  \item 
    What are the limits of cross-domain generalizability for models trained on semi-structured \textit{StVO} when applied to foreign driving regulations with varying degrees of overlap?
  \item 
        How effectively does \textit{StVO} legal
        knowledge transfer to other law-related tasks beyond
        traffic regulations?
\end{enumerate}

\subsection{Datasets}
\label{sec:datasets}
\paragraph{German Road Traffic Regulations (\textit{StVO}).} 
The German Road Traffic Regulations (\textit{Straßenverkehrsordnung};\textit{StVO}) \citep{bmdv2013stvo} form the core corpus for our \(KG\) pre-training approach. The \textit{StVO} is a highly semi-structured, self-contained statute  governing public road traffic and operations in Germany. Rather than treating it as flat text, we exploit its hierarchical layout to construct a formal \(KG\) (see the visual topology in Appendix~\ref{sec:appendix_kg_viz}).

\paragraph{Synthetic \textit{StVO} QA Benchmark.}
Existing legal QA benchmarks~\citep{Guha2023LegalBenchAC} focus on case law, and lack citation grounding.
We use LLM-based
generation~\citep{AugGPTarticle} to produce 12,554 labeled balanced
QA pairs.
For each \textit{StVO} sentence $s_i$, we prompt
Mistral-7B-Instruct~\citep{jiang2023mistral7b} and
Llama-3.1-8B~\citep{grattafiori2024llama} to generate
open-ended questions with paired correct ($y_{\text{pos}}$,
source-grounded) and incorrect ($y_{\text{neg}}$, plausible but
contradictory) answers. Dataset statistics and representative example are provided in Table~\ref{tab:synthetic_dataset_stats} and Box~\ref{box:example_QA_pair}, respectively.
To ensure legal reliability, each generated pair is scored by a composite
quality metric $Q \in [0,1]$, and only pairs with $Q \geq 0.6$ are retained.
$Q$ starts at $1.0$ and accumulates penalties across five checks:
(1)~structural and field validity, enforcing required fields and length bounds
(questions 15--200, answers 20--500 words);
(2)~rejection of single-word yes/no items that carry no reasoning;
(3)~presence of domain terminology (e.g., \textit{Vorfahrt}, right of way);
(4)~lexical-overlap control, penalising question--answer overlap above 70\%
while requiring at least 10\% source grounding; and
(5)~placeholder detection.
The thresholds are precision-oriented heuristics fixed a priori from manual
inspection of generated pairs against the statutes, independently of downstream
results; $Q \geq 0.6$ balances dataset precision against the scale needed for
pre-training. 
Following \citep{lewis-etal-2021-paq}, we prevent data leakage via \textbf{paragraph-level splitting}. By reserving 17\% of paragraph identifiers exclusively for testing, we ensure the model cannot exploit content memorization. This yields a robust split of \textbf{10,245 train / 2,309 test} pairs with zero paragraph overlap (Table~\ref{tab:synthetic_dataset_stats}).
\vspace{-0.5em}
\begin{table}[H]
\centering
\renewcommand{\arraystretch}{1.1} 
\small
\begin{tabular}{lrr}
\toprule
\textbf{Statistic} & \textbf{Train} & \textbf{Test} \\
\midrule
QA pairs            & 10,245          & 2,309          \\
Unique paragraphs   & 49              & 10             \\
Positive samples    & 50.1\%          & 49.4\%         \\
Negative samples    & 49.9\%          & 50.6\%         \\
\bottomrule
\end{tabular}
\caption{Synthetic \textit{StVO} QA dataset statistics.}
\label{tab:synthetic_dataset_stats}
\end{table}

\begin{examplebox}[label={box:example_QA_pair}]{Generated QA Pair --- \S~8 Right of way --- \textit{StVO}}
Given the legal sentence from paragraph \S~8 (\textit{StVO}): \textit{``At intersections without traffic signs or lights, vehicles coming from the right have right-of-way''}, the LLM generates:
\smallskip
\noindent\textbf{Correct pair} ($y_{\text{pos}}$):
\begin{itemize}[nosep, leftmargin=1.5em, label=\textbullet]
    \item Q: \textit{``Which vehicle has priority at an unmarked intersection?''}
    \item A: \textit{``At intersections without traffic signs or lights, vehicles approaching from the right have right-of-way.''}
\end{itemize}
\smallskip
\noindent\textbf{Incorrect pair} ($y_{\text{neg}}$):
\begin{itemize}[nosep, leftmargin=1.5em, label=\textbullet]
    \item Q: \textit{``Who must yield at an intersection without signals?''}
    \item A: \textit{``The vehicle traveling at lower speed must always yield to faster-moving traffic.''} [Incorrect: contradicts the right-hand priority rule]
\end{itemize}
\end{examplebox}

\paragraph{In-Domain Traffic Regulation Benchmarks.}
To assess out-of-domain generalization, we utilize two open-source road safety datasets: 890 Irish theory questions \citep{adamisntdead2024theory} and 1,954 Austrian \textit{ÖAMTC Führerschein-Test} questions \citep{dsumer2024driving}. We cast each multiple-choice item as a binary correctness judgment against individual options. After excluding image-dependent samples to ensure a purely linguistic evaluation, the resulting benchmarks comprise 2,500 test samples (from the remaining textual 625 questions) for the Irish set and 3,852 samples (from the remaining textual 963 questions) for the Austrian set.\paragraph{LexGLUE Benchmark.}\citet{chalkidis-etal-2022-lexglue} provides seven English legal NLP tasks, enabling comparison with SOTA models. By benchmarking on LexGLUE, we probe the extent to which our \(KG\)-guided pre-training approach generalizes beyond statutory context.

\subsection{Models and Training}
\label{subsec:models_training}
We focus on encoder-only architectures, which remain the standard choice for legal-understanding tasks of this kind. Our objective is a binary correctness
judgment over a (question, answer, paragraph) triple. Bidirectional encoders suit this discriminative classification setup better than generative models. Extending the objectives to decoder-only models is left to future work.

We evaluate nine transformer-based encoders: \textbf{1) five legal-domain}
(\texttt{bert-base-eurlex}, \texttt{bert-base-echr}, \texttt{bert-base-contracts},
\texttt{custom-legalbert}~\citep{chalkidis2020legalbert}, and
\texttt{Italian-Legal-BERT}~\citep{Licari2022ITALIANLEGALBERTAP});
\textbf{2) three general-purpose} (\texttt{google-bert-base}~\citep{devlin2019bert},
\texttt{legal-bert-base}, \texttt{legal-bert-small}~\citep{chalkidis2020legalbert});
and \textbf{3) one non-Latin cross-lingual} (\texttt{Legal-heBERT}~\citep{legalHebert22}).

We explore \textbf{36 $\lambda$ configurations} over $\{0.0, 0.1, \ldots, 0.9\}$ per task under the unit-sum constraint ($\lambda_\text{NSP} + \lambda_\text{CLU} + \lambda_\text{CLS} = 1$), yielding \textbf{324 experimental runs} in total ($9~\text{models} \times 36~\text{configs}$). 
The full configuration grid is described in Appendix \ref{appx:hyperparameters}.

\subsection{Evaluation Metrics}
To provide a comprehensive assessment of the model performance, we report standard classification metrics: Macro-$F_1$ ($m\text{-}F_1$), Micro-$F_1$ ($\mu\text{-}F_1$) \citep{buckland1994relationship}, Precision ($P$) and Recall ($R$) \citep{powers2020evaluation}. To account for potential class imbalances in our synthesized data, we include Balanced Accuracy \citep{brodersen2010balanced}. Finally, we report the absolute gain in $m\text{-}F_1$ over the respective base models to quantify the direct impact of our training approach.

\section{Results and Discussion}
\label{sec:results}
\subsection{Impact of KG-aware Pre-Training on \textit{StVO} Legal Reasoning}
\label{subsec:results_stvo}
This part evaluates how incorporating structured \(KG\) training directly
influences the legal reasoning capabilities of PLMs. Multi-task pre-training yields large, consistent gains, but its magnitude depends sharply on the base model's domain and capacity (Table~\ref{tab:stvo_synthetic_data_results}).

All five legal-domain encoders exceed $0.8$ $m\text{-}F_1$ after pre-training, and the gain is largely decoupled from base
performance. \texttt{bert-base-eurlex} rises from the weakest base ($0.338$)
to $0.862$ (the largest absolute gain, $+0.524$), while \texttt{bert-base-echr}
reaches the overall peak ($0.864$). This decoupling indicates the improvement
stems from the KG-guided objectives rather than from a strong starting point. Counterintuitively, the general-purpose \texttt{google-bert-base} gains more ($+0.205$, to $0.630$) than the broad, multi-field \texttt{legal-bert-base} ($+0.114$ to $0.569$). We attribute this to \emph{domain-specific interference} where multiple legal English tasks conflict with the \textit{StVO} domain. On the other hand, the single-subfield specialists like echr and eurlex align with it cleanly.

The two weakest cases fail for distinct reasons. \texttt{legal-bert-small}
plateaus at $0.537$ ($+0.050$), consistent with a capacity bottleneck that we
examine across configurations in Section \ref{subsec:tasksynergy}.
\texttt{Legal-heBERT} instead suffers from a severe precision-recall imbalance ($P=1.000$, $R=0.096$), 
reflecting a fundamental linguistic and structural incompatibility between the Hebrew pre-training and the English target. This shows up as monotonic decay under task scaling (Figure~\ref{fig:task_scaling_analysis}). Consequently, \texttt{Legal-heBERT} cannot exploit the \(KG\) training objectives at all, and any gains it shows reflect the English CLS head alone rather than \(KG\) pre-training.

Overall, legal-domain models outperform general-purpose ones after \(KG\)
pre-training, indicating that domain knowledge \emph{synergizes} with the
objectives rather than substituting for them. This is subject to sufficient capacity and language compatibility.

\begin{table*}[t]
\centering
\footnotesize
\setlength{\tabcolsep}{9pt}
\begin{tabular}{lc|cccccc|cc}
\toprule
& & \multicolumn{6}{c|}{\textit{Multi-task Pre-training on \textit{StVO} (Zero-shot evaluations)}}
  & \multicolumn{2}{c}{\textit{Fine-tuning}} \\
\cmidrule(lr){3-8}\cmidrule(lr){9-10}
\textbf{Model}
  & \textbf{Base}
  & \textbf{PT}
  & \textbf{PT}
  & \textbf{PT}
  & \textbf{PT}
  & \textbf{PT}
  & \textbf{PT}
  & \textbf{FT}
  & \textbf{FT} \\
  & $m\text{-}F_1$
  & $m\text{-}F_1$
  & Gain
  & $\mu\text{-}F_1$
  & Bal.~Acc.
  & P & R
  & $m\text{-}F_1$
  & Gain \\
\midrule
\multicolumn{10}{l}{\textit{Top Performers ($m\text{-}F_1>0.8$)}} \\
bert-base-echr       & 0.566 & \textbf{0.864} & $+$0.298 & 0.865 & 0.864 & 0.937 & 0.779 & 0.912 & $+$0.048 \\
bert-base-eurlex     & 0.338 & \textbf{0.862} & $+$0.524 & 0.863 & 0.864 & 0.812 & 0.939 & 0.909 & $+$0.047 \\
bert-base-contracts  & 0.475 & \textbf{0.845} & $+$0.370 & 0.846 & 0.848 & 0.784 & 0.950 & 0.905 & $+$0.060 \\
custom-legalbert     & 0.467 & \textbf{0.820} & $+$0.353 & 0.820 & 0.820 & 0.812 & 0.828 & 0.900 & $+$0.080 \\
Italian-Legal-BERT   & 0.514 & \textbf{0.813} & $+$0.299 & 0.816 & 0.815 & 0.905 & 0.701 & 0.893 & $+$0.080 \\
\midrule
\multicolumn{10}{l}{\textit{Moderate Performers}} \\
google-bert-base     & 0.425 & \textbf{0.630} & $+$0.205 & 0.662 & 0.666 & 0.597 & 0.970 & 0.919 & $+$0.289 \\
legal-bert-base      & 0.456 & \textbf{0.569} & $+$0.114 & 0.613 & 0.617 & 0.565 & 0.945 & 0.888 & $+$0.319 \\
\midrule
\multicolumn{10}{l}{\textit{Limited Improvement}} \\
legal-bert-small     & 0.486 & \textbf{0.537} & $+$0.050 & 0.544 & 0.545 & 0.530 & 0.671 & 0.892 & $+$0.355 \\
Legal-heBERT         & 0.336 & \textbf{0.435} & $+$0.099 & 0.554 & 0.548 & 1.000 & 0.096 & 0.870 & $+$0.435 \\
\bottomrule
\end{tabular}
\caption{Model performance on the synthetic \textit{StVO}
Q\&A benchmark. \textbf{Base}: $m\text{-}F_1$ with no \textit{StVO}
knowledge; \textbf{PT}: after multi-task pre-training; \textbf{FT}:
after fine-tuning on a few downstream samples from the synthetic \textit{StVO} Q\&A benchmark. PT and FT Gain report absolute $m\text{-}F_1$ improvement over Base and PT scores, respectively.}
\label{tab:stvo_synthetic_data_results}
\end{table*}

\subsection{Cross-Lingual Transfer (RQ1)}
\label{subsec:results_crosslingual}
To isolate cross-lingual transfer of legal knowledge, we train the
NSP and CLU objectives exclusively on non-translated \emph{German}
\textit{StVO} text, while the CLS head is trained on a subset of the
\emph{English} synthesized QA pairs. We then evaluate zero-shot on the English
test set. Because NSP and CLU never see English, any zero-shot gain reflects German-learned structure surfacing through the English classification head (Table~\ref{tab:cross_lingual_transfer}).

Five models achieve positive zero-shot gains ($+0.094$ to $+0.499$). The knowledge learned from German NSP+CLU is recoverable in English. \texttt{bert-base-eurlex} transfers most strongly ($0.338\!\to\!0.837$, $+0.499$) and barely improves with further fine-tuning ($+0.032$), indicating that the legal knowledge already injected through German NSP and CLU is sufficient on its own, leaving little for target-language fine-tuning to add. Three models transfer poorly zero-shot (\texttt{bert-base-echr},
\texttt{custom-legalbert}, \texttt{legal-bert-small}; down to $-0.148$) but
recover fully when fine-tuned, with \texttt{legal-bert-small} posting the
largest FT gain ($+0.532$, to $0.871$). All models converge to a narrow band
($0.862$--$0.878$), indicating German pre-training encodes the required
legal knowledge regardless of transfer mode.

\begin{table*}[t]
\centering
\footnotesize
\setlength{\tabcolsep}{9pt}
\begin{tabular}{lc|cccccc|cc}
\toprule
& & \multicolumn{6}{c|}{\textit{Multi-task Pre-training on StVO (Zero-shot evaluations)}}
  & \multicolumn{2}{c}{\textit{Fine-tuning}} \\
\cmidrule(lr){3-8}\cmidrule(lr){9-10}
\textbf{Model}
  & \textbf{Base}
  & \textbf{PT}
  & \textbf{PT}
  & \textbf{PT}
  & \textbf{PT}
  & \textbf{PT}
  & \textbf{PT}
  & \textbf{FT}
  & \textbf{FT} \\
  & $m\text{-}F_1$
  & $m\text{-}F_1$
  & Gain
  & $\mu\text{-}F_1$
  & Bal.~Acc.
  & P & R
  & $m\text{-}F_1$
  & Gain \\
\midrule
\multicolumn{10}{l}{\textit{Successful Cross-Lingual Transfer}} \\
bert-base-eurlex     & 0.338 & \textbf{0.837} & $+$0.499 & 0.837 & 0.837 & 0.832 & 0.839 & 0.869 & $+$0.032 \\
google-bert-base     & 0.425 & \textbf{0.773} & $+$0.347 & 0.775 & 0.777 & 0.719 & 0.893 & 0.871 & $+$0.099 \\
bert-base-contracts  & 0.475 & \textbf{0.694} & $+$0.219 & 0.701 & 0.699 & 0.776 & 0.555 & 0.878 & $+$0.184 \\
legal-bert-base      & 0.456 & \textbf{0.599} & $+$0.144 & 0.628 & 0.632 & 0.579 & 0.909 & 0.876 & $+$0.276 \\
Italian-Legal-BERT   & 0.514 & \textbf{0.608} & $+$0.094 & 0.635 & 0.632 & 0.764 & 0.377 & 0.876 & $+$0.268 \\
\midrule
\multicolumn{10}{l}{\textit{Failed Zero-Shot, Recovered via Fine-Tuning}} \\
bert-base-echr       & \textbf{0.566} & 0.563 & $-$0.003 & 0.606 & 0.610 & 0.561 & 0.933 & 0.867 & $+$0.304 \\
custom-legalbert     & \textbf{0.467} & 0.377 & $\mathbf{-}$0.090 & 0.517 & 0.511 & 0.671 & 0.043 & 0.862 & $\mathbf{+}$0.485 \\
legal-bert-small     & \textbf{0.486} & 0.339 & $\mathbf{-}$0.148 & 0.411 & 0.415 & 0.443 & 0.749 & 0.871 & \textbf{$\mathbf{+}$0.532} \\
\bottomrule
\end{tabular}
\caption{The objectives (NSP+CLU) are pre-trained on non-translated German \textit{StVO}. Models are evaluated on the synthetic \textit{StVO} Q\&A benchmark. \textbf{Base}: $m\text{-}F_1$ before
pre-training. \textbf{PT} columns report zero-shot performance after pre-training. \textbf{PT Gain} $=$ PT $m\text{-}F_1 -$ Base. \textbf{FT} columns report performance after fine-tuning on a few downstream samples from the synthetic \textit{StVO} Q\&A. \textbf{FT Gain} $=$ FT $-$ PT. Bold marks the higher of Base and PT $m\text{-}F_1$ per model.}
\label{tab:cross_lingual_transfer}
\end{table*}

\subsection{In-Domain Traffic-Law Transfer (RQ2)}
\label{subsec:results_ood}

\begin{figure}[t]
\centering
\begin{tikzpicture}
\begin{axis}[
    width=\columnwidth,
    height=5.5cm,
    ybar=3pt,
    bar width=6pt,
    ymajorgrids=true,
    grid style={dotted,gray!40},
    ylabel={$\Delta\,m$-$F_1$ over Base},
    ylabel style={font=\scriptsize},
    ymin=-0.07, ymax=0.18,
    ytick={-0.05,0.00,0.05,0.10,0.15},
    yticklabel style={font=\scriptsize},
    xtick=data,
    xticklabels={
        Legal-heBERT,
        google-bert-base,
        custom-legalbert,
        legal-bert-small,
        bert-base-echr,
        bert-base-contracts,
        Italian-Legal-BERT,
        bert-base-eurlex,
        legal-bert-base
    },
    xticklabel style={
        font=\tiny,
        rotate=35,
        anchor=north east
    },
    legend style={
        at={(0.5,1.02)},
        anchor=south,
        legend columns=2,
        font=\scriptsize,
        column sep=6pt,
        draw=none
    },
    enlarge x limits=0.08,
    clip=false,
]
\addplot[fill=blue!60, draw=blue!80]
coordinates {
    (1, 0.0341)(2, 0.1582)(3, 0.1263)(4, 0.0694)
    (5, 0.1231)(6, 0.1196)(7, 0.0845)(8, 0.1145)(9, 0.0753)
};
\addplot[fill=orange!70, draw=orange!90]
coordinates {
    (1,-0.0489)(2, 0.0637)(3, 0.0363)(4, 0.0525)
    (5, 0.0821)(6, 0.0016)(7, 0.0615)(8, 0.0014)(9, 0.0603)
};
\legend{Austrian ($+$StVO), Irish ($+$StVO)}
\draw[black!50, thin] (axis cs:0.4,0) -- (axis cs:9.6,0);
\end{axis}
\end{tikzpicture}
\caption{$m$-$F_1$ gain of StVO multi-task pre-training over
each base model on Austrian and Irish zero-shot driving exams. 
Austrian gains are consistently larger, reflecting
greater regulatory overlap with German \textit{StVO}.
Legal-heBERT is the only model showing negative transfer on
Irish ($-$0.049).}
\label{fig:transfer_gains}
\end{figure}
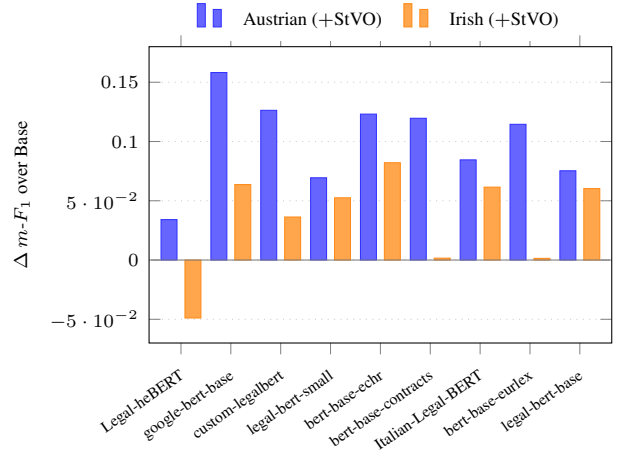

In Figure~\ref{fig:transfer_gains} all tested models improve zero-shot on the
Austrian benchmark ($+0.041$ to $+0.158$), confirming transfer within the same
regulatory family. Irish gains are smaller and less consistent ($+0.002$ to $+0.082$), and \texttt{Legal-heBERT} is the only model with negative transfer ($-0.049$), consistent with its limitations in Section \ref{subsec:results_stvo}. Austrian gains exceed Irish gains for nearly every model because the Austrian \textit{ÖAMTC} regulations share greater regulatory overlap with \textit{StVO}.

\textit{StVO} pre-training generalizes within the European traffic-regulation family, but the gains are modulated more by linguistic distance (Table~\ref{tab:cross_lingual_transfer}) than by traffic law distance (Figure~\ref{fig:transfer_gains}). Together, these results indicate that cross-linguistic distance is a stronger barrier to transfer than cross-regulatory distance within European traffic law.

\subsection{Out-of-Domain Legal Transfer (RQ3)}
\label{subsec:results_lexglue}

Here, we probe whether \textit{StVO} pre-training transfers beyond traffic law to
seven English legal tasks (Table~\ref{tab:lexglue_direct_comparison}). Despite
training exclusively on German traffic regulations, our models improve over the published ~\citep{chalkidis-etal-2022-lexglue} baselines on four of the seven tasks—ECtHR-A, ECtHR-B, UNFAIR-ToS, and CaseHOLD—indicating that the approach encodes legal reasoning transferable beyond the \textit{StVO} domain. See Appendix~\ref{app:lexglue} for the complete task evaluations.



\begin{table}[t]
\centering
\setlength{\tabcolsep}{4pt}
\renewcommand{\arraystretch}{1.1}
\resizebox{\columnwidth}{!}{%
\begin{tabular}{@{}l cc | cc | cc | c@{}}
\toprule
\multirow{2}{*}{\textbf{Model}}
  & \multicolumn{2}{c|}{\textbf{ECtHR-A}}
  & \multicolumn{2}{c|}{\textbf{ECtHR-B}}
  & \multicolumn{2}{c|}{\textbf{UNFAIR-ToS}}
  & \textbf{CaseHOLD} \\
\cmidrule(lr){2-3}\cmidrule(lr){4-5}\cmidrule(lr){6-7}\cmidrule(lr){8-8}
  & $\mu\text{-}F_1$ & $m\text{-}F_1$ & $\mu\text{-}F_1$ & $m\text{-}F_1$ & $\mu\text{-}F_1$ & $m\text{-}F_1$ & $F_1$ \\
\midrule
\multicolumn{8}{@{}l}{\textit{SOTA Published Baselines}} \\
TFIDF-SVM             & 62.6 & 48.9 & 73.0 & 63.8 & 94.7 & 75.0 & 22.4 \\
BERT                  & 71.2 & 63.6 & 79.7 & 73.4 & 95.6 & 81.3 & 70.8 \\
RoBERTa               & 69.2 & 59.0 & 77.3 & 68.9 & 95.2 & 79.2 & 71.4 \\
DeBERTa               & 70.0 & 60.8 & 78.8 & 71.0 & 95.5 & 80.3 & 72.6 \\
Longformer            & 69.9 & 64.7 & 79.4 & 71.7 & 95.5 & 80.9 & 71.9 \\
Legal-BERT            & 70.0 & 64.0 & 80.4 & \textbf{74.7} & 96.0 & 83.0 & 75.3 \\
CaseLaw-BERT          & 69.8 & 62.9 & 78.8 & 70.3 & 96.0 & 82.3 & 75.4 \\
\midrule
\multicolumn{8}{@{}l}{\textit{Ours (\textit{StVO} multi-task)}} \\
BERT-EurLex+StVO      & 70.8 & \textbf{65.7}$^\dagger$ & 79.7 & 73.8 & 95.7 & 79.4 & 71.5 \\
BERT-ECHR+StVO        & \textbf{72.7}$^\dagger$ & 61.2 & \textbf{81.0}$^\dagger$ & 73.9 & 95.8 & 82.5 & 71.9 \\
Custom-LegalBERT+StVO & 71.2 & 61.6 & 78.6 & 70.6 & \textbf{96.8}$^\dagger$ & \textbf{86.0}$^\dagger$ & \textbf{77.0}$^\dagger$ \\
\bottomrule
\end{tabular}}
\caption{LexGLUE results on the four tasks where \textit{StVO}
pre-training improves over the published baselines ~\citep{chalkidis-etal-2022-lexglue}. (full seven-task results in
Appendix~\ref{app:lexglue}). Scores are $F_1$ (\%): macro ($m$) and micro
($\mu$). \textbf{Bold} marks the best score per column; $\dagger$ marks the improvement over the baselines of~\citet{chalkidis-etal-2022-lexglue}. CaseHOLD is
reported as a single $F_1$ ($\mu{=}m$).}
\label{tab:lexglue_direct_comparison}
\end{table}

\subsection{Task synergy and interference}
\label{subsec:tasksynergy}
\begin{figure}[H]
\centering
\begin{tikzpicture}
\begin{axis}[
    title={\small\bfseries Task Scaling: Synergies and
           Interference},
    xlabel={\small Number of Pre-training Tasks},
    ylabel={\small F$_1$-Macro Score},
    xmin=0.8, xmax=3.2,
    ymin=0.30, ymax=1.0,
    xtick={1,2,3},
    xticklabels={\small Single,\small Two,\small All Three},
    ymajorgrids=true,
    grid style={line width=.1pt, draw=gray!20},
    width=\columnwidth,
    height=5.2cm,
    legend style={
        at={(0.5,-0.30)}, anchor=north,
        font=\tiny, legend columns=2,
        cells={align=left}, column sep=4pt, row sep=0pt,
    },
    legend cell align={left},
    axis lines=left, clip=false
]
\pgfplotsset{thick}
\addplot[color=red!70!black, mark=square*, mark size=2pt]
  coordinates {(1,0.587)(2,0.826)(3,0.864)};
\addlegendentry{bert-base-echr ($\star$)}
\addplot[color=blue!70!black, mark=*, mark size=2pt]
  coordinates {(1,0.675)(2,0.833)(3,0.862)};
\addlegendentry{bert-base-eurlex}
\addplot[color=green!50!black, mark=triangle*, mark size=2pt]
  coordinates {(1,0.522)(2,0.684)(3,0.820)};
\addlegendentry{custom-legalbert}
\addplot[color=orange!90!black, mark=diamond*, mark size=2pt]
  coordinates {(1,0.537)(2,0.806)(3,0.813)};
\addlegendentry{Italian-Legal-BERT}
\addplot[color=purple!80!black, mark=pentagon*,
         mark size=2pt, dashed]
  coordinates {(1,0.643)(2,0.845)(3,0.797)};
\addlegendentry{bert-base-contracts ($\dagger$)}
\addplot[color=gray!60!black, mark=star,
         mark size=2pt, dashed]
  coordinates {(1,0.480)(2,0.630)(3,0.620)};
\addlegendentry{google-bert-base ($\dagger$)}
\addplot[color=brown!80!black, mark=otimes*, mark size=2pt]
  coordinates {(1,0.344)(2,0.510)(3,0.569)};
\addlegendentry{legal-bert-base}
\addplot[color=teal!80!black, mark=oplus*, mark size=2pt]
  coordinates {(1,0.344)(2,0.504)(3,0.537)};
\addlegendentry{legal-bert-small}
\addplot[color=magenta!80!black, mark=x, mark size=3pt,
         dashed, line width=1pt]
  coordinates {(1,0.435)(2,0.401)(3,0.350)};
\addlegendentry{Legal-heBERT ($\ddagger$)}
\end{axis}
\end{tikzpicture}
\caption{Task scaling across nine models. \textbf{bert-base-echr
($\star$)} peaks at full-task synergy~(0.864). ($\dagger$)~models
show negative CLU interference when added to NSP+CLS.
\textbf{Legal-heBERT ($\ddagger$)} degrades monotonically.}
\label{fig:task_scaling_analysis}
\end{figure}
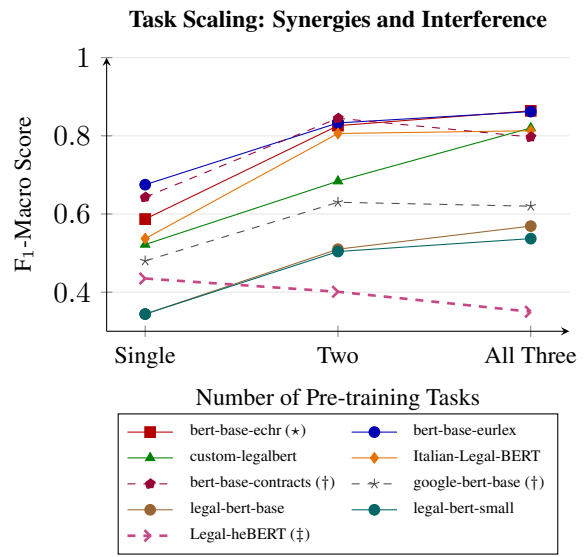

We ablate the three objectives across all models and 36 $\lambda$ configurations to isolate each one's contribution, as shown in Figure~\ref{fig:task_scaling_analysis}. The full NSP+CLU+CLS configuration outperforms all partial task subsets for 7 of the 9 evaluated models. For \texttt{bert-base-echr}, the integration of all three objectives achieves the peak performance, elevating $m\text{-}F_1$ from 0.587 (CLS only) to 0.864 (ALL), a gain of $+$0.277 over the strongest standalone baseline. Aggregated results identify CLS as the dominant objective. Its removal precipitates a mean $m\text{-}F_1$ drop of 0.277. \texttt{bert-base-contracts} in Figure~\ref{fig:task_scaling_analysis} presents a notable exception to the observed synergy: the NSP+CLS configuration (0.845) outperforms the full ALL suite (0.797). We attribute this performance degradation to two primary factors: (a)~a semantic domain mismatch between the legal entities of the \textit{StVO} and the established legal primitives in US contract law, and (b)~gradient interference within the CLU objective, where random negative sampling potentially pairs semantically related paragraphs as negatives, thereby introducing some noise that might destabilize the optimization of this specific architecture. The performance degradation observed for \texttt{legal-bert-small} in Figure~\ref{fig:task_scaling_analysis} suggests a capacity bottleneck, where the model's limited parameter space fails to capture the synergy between complex training objectives. The results conclude that the full configuration or dual-objective subset outperforms any single objective. The CLS task acts as the dominant contributor because it aligns directly with the downstream task structure.

\section{Conclusion and Future Work}
\label{sec:conclusion}

This paper presented a knowledge graph-guided multi-task pre-training approach
for \textit{StVO} legal question answering, encoding structured regulatory
knowledge through three complementary objectives. Our main finding is that
explicitly encoding legal structure via knowledge-graph-guided pre-training
yields compounding benefits beyond domain-adaptive single-objective training.

Across the evaluated models, legal-domain models consistently outperform
general-purpose ones, confirming that legal-domain knowledge leverages 
the objectives rather than being replaced by them. Model capacity and linguistic distance can nonetheless override this advantage. Our ablation identifies CLS as the dominant objective, with NSP and CLU serving as complementary structural tasks. The full configuration benefits most models, and the exceptions show that this synergy depends on semantic compatibility between the knowledge graph and the base model.
The approach also transfers beyond its source domain. German structural
pre-training transfers cross-lingually to English. It further yields gains on Austrian and Irish driving regulations and improves over established LexGLUE baselines on several English legal benchmarks. Together, these results indicate that knowledge-graph-guided pre-training on a single regulatory domain can bridge to structurally similar legal reasoning tasks across languages.

Future work will replace random CLU negative sampling with hard-negative sampling and extend the approach to other European regulatory corpora.

\section*{Data and Code Availability}
To support reproducibility, we release the full pipeline code, the synthetic StVO QA dataset, the parsed knowledge graph, the TransE embeddings, and the segmenter patterns. The pre-training and modeling code is available at \url{https://github.com/ibrahimssd/stvo-llm-driver}, and the QA dataset and knowledge-graph resources at \url{https://github.com/ibrahimssd/stvo-legal-qa}. All hyperparameters are specified in Appendix~\ref{appx:hyperparameters} (seed 42, single H100).

\section*{Limitations}
\paragraph{Domain and language scope.}
Our approach is evaluated only on German traffic regulations (\textit{StVO}).
Generalization to other legal systems, languages, and regulatory domains
remains to be validated empirically.

\paragraph{Synthetic \textit{StVO} Benchmark.}
The 12{,}554 QA pairs are LLM-generated and may not reflect authentic
examination distributions, particularly with respect to ambiguity and
multi-hop reasoning. Because the training and evaluation pairs share the same
generators, the reported gains should be read as evidence on synthetic legal QA
rather than on authentic exams.

\paragraph{CLU negative sampling.}
Random negative sampling can pair semantically related paragraphs as negatives,
introducing interference for out-of-domain bases such as
\texttt{bert-base-contracts}. Possible alternatives such as hard-negative sampling is 
left to future work.

\paragraph{Knowledge-graph dependence.}
The KG is parsed from an LLM-translated English of the \textit{StVO},
and the CLU and CLS objectives depend on TransE embeddings derived from it. This
pipeline introduces translation and embedding noise that we do not quantify. The CLS attention also scales over the full set of paragraph embeddings. This is tractable for the 59-paragraph StVO graph, but its cost grows linearly with the number of paragraphs and is untested for larger legal codes. Because the paragraph embeddings are computed once per domain and remain fixed during pre-training, this bottleneck can be mitigated by replacing the dense attention with approximate nearest-neighbor retrieval (e.g., FAISS, ~\citealp{johnson2019billion}) over the embedding library so that each query attends only to the top-$k$ relevant paragraphs rather than the full set. We leave a systematic study of this scaling behavior to future work.


\section*{Ethical Considerations}

The models presented here are research prototypes trained on synthetic
question-answer pairs derived from the German \textit{StVO}. They are not
intended to provide legally binding advice, and their outputs must not
substitute for consultation with a qualified legal professional. Deployment in
safety-critical settings such as  automated driver assessment without
human oversight is strongly discouraged. The pre-training and evaluation corpora are LLM-generated and passed through a quality-filtering pipeline. Filtering does not guarantee correctness. The
synthetic QA pairs may still contain factual inaccuracies or misrepresent legal
provisions. Because no held-out subset was validated by human legal experts,
the true legal reasoning depth of the data remains uncertain. The models are trained exclusively on German traffic law. Cross-traffic transfer to Austrian and Irish driving regulations is evaluated zero-shot and
shows variable performance. Users should not apply these models outside the German traffic-law 
context without further validation. The base encoders are pre-trained on existing legal corpora such as EUR-LEX and may encode biases present in those sources. Continued pre-training on \textit{StVO} does not remove such biases, and their downstream effects on legal reasoning have not been systematically audited in this work.

\bibliography{references.bib}

\appendix

\section{Knowledge Graph Topology and Visualization}
\label{sec:appendix_kg_viz}
\begin{figure}[t]
    \centering
    \includegraphics[width=0.49\textwidth]{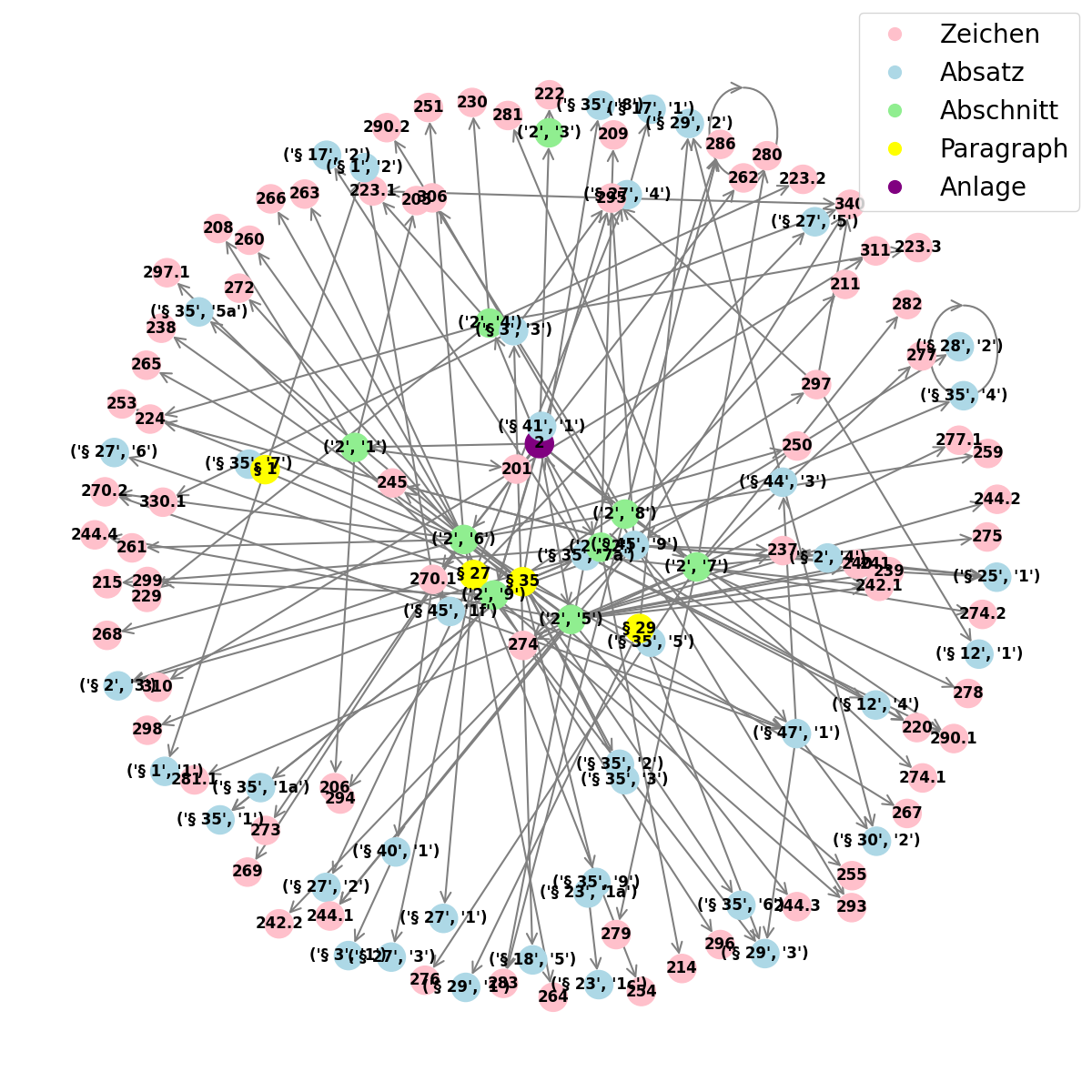}
    \caption{Partial visual topology of the generated \textit{StVO} knowledge graph (554 nodes, 945 edges). Every node has tuples that correspond to its unique IDs, and the node content incorporates the related legal knowledge}
    \label{fig:appendix_kg}
\end{figure}

Figure~\ref{fig:appendix_kg} illustrates a portion of the knowledge graph (\(KG\)) that was created using the suggested parsing approach \citep{Siddig2025_1260} and derived from the Straßenverkehrs-Ordnung (\textit{StVO}). This section presents the topological layout of the extracted knowledge base to provide structural transparency regarding the legal ontologies used during multi-task pre-training. The German \textit{Straßenverkehrsordnung} (\textit{StVO}) corpus was translated and systematically parsed into a legal knowledge graph (\(KG\)) structure. The resulting network comprises 554 discrete nodes and 945 directed edges. Nodes represent specific regulatory elements, such as paragraphs (\S). Edges define the directional dependencies, nested hierarchies, and cross-references that bind these paragraphs together. 

\section{Hyperparameter Configuration}
\label{appx:hyperparameters}

This appendix reports the complete hyperparameter settings for the training stages of our pipeline: (B.1) TransE knowledge graph embedding and (B.2) \(KG\)-guided multi-task pre-training. All the aforementioned stages use seed \texttt{42} and a single NVIDIA H100 with 80\,GB and CUDA 12.1.

\subsection{TransE Embedding Training}
\label{appx:transe_hparams}

We train TransE on the German \textit{StVO} triple set with 554 nodes, 945 edges, and 5 relations to produce the 500-dimensional paragraph embeddings ($\mathbf{e}_p$) used as CLU soft targets (Section \ref{subsec:multitask_pretraining}). The model is optimized with a margin-ranking loss over uniformly corrupted head/tail negatives. The best checkpoint is selected by validation loss.

\begin{table}[t]
\centering
\small
\begin{tabular}{@{}ll@{}}
\toprule
\textbf{Hyperparameter} & \textbf{Value} \\
\midrule
Scoring function          & $-\lVert \mathbf{e}_h + \mathbf{r} - \mathbf{e}_t \rVert_2$ \\
Embedding dimension $d$   & 500 \\
Distance norm $p$         & 2 (L2) \\
Margin $\gamma$           & 1.0 \\
Loss                      & Margin ranking \\
Negative sampling         & Uniform (head + tail) \\
Optimizer                 & Adam \\
Learning rate             & $1\times10^{-3}$ \\
Weight decay              & $1\times10^{-6}$ \\
LR scheduler              & ReduceLROnPlateau \\
$L_2$ regularization      & 0.01 \\
Dropout                   & 0.1 \\
Batch size                & 256 \\
Epochs                    & 50 \\
Gradient clipping         & 1.0 (max-norm) \\
Initialization            & Xavier uniform, renorm to 1 \\
Train / val / test split  & 80\,/\,10\,/\,10 \\
Model selection           & Best validation loss \\
\bottomrule
\end{tabular}
\caption{TransE hyperparameters for the \textit{StVO} \(KG\) paragraph embedding.}
\label{tab:transe_hparams}
\end{table}

\subsection{KG-Guided Multi-Task Pre-Training}
\label{appx:mtl_hparams}

All base encoders use the masked-modeling configuration, combining NSP with CLU and CLS. The three loss weights satisfy $\lambda_{\text{NSP}}+\lambda_{\text{CLU}}+\lambda_{\text{CLS}}=1$;
we sweep the grid $\{0.0,0.1,\dots,0.9\}$ subject to this unit-sum
constraint. $\lambda_{\text{CLU}}$ is determined as $1-(\lambda_{\text{NSP}}+\lambda_{\text{CLS}})$, yielding 36 valid configurations per model (Section \ref{subsec:models_training}).

\begin{table}[t]
\centering
\small
\begin{tabular}{@{}ll@{}}
\toprule
\textbf{Hyperparameter} & \textbf{Value} \\
\midrule
Modeling type             & Masked (NSP core) \\
KG embedding source       & TransE, $d=500$ \\
Optimizer                 & AdamW \\
Learning rate             & $5\times10^{-5}$ \\
LR scheduler              & Linear w/ warmup \\
warmup ratio        & 0.10 \\
Epochs                    & 20 \\
Batch size (per task)     & 16 (NSP, CLU, CLS) \\
Gradient accumulation     & 2 \\
Effective batch size      & 32 \\
Max sequence length       & 128 (NSP / CLU / CLS) \\
Max segments per legal sentence      & 10 \\
CLS label space           & 2 (correct / incorrect) \\
$\lambda$ grid            & $\{0.0,\dots,0.9\}$, unit-sum \\
Configurations per model  & 36 \\
\bottomrule
\end{tabular}
\caption{Multi-task pre-training hyperparameters, shared across all base encoders.}
\label{tab:mtl_hparams}
\end{table}



\section{Full Ablation Results}
\label{app:full_results}

\begin{figure*}[t]
\centering
\begin{subfigure}[b]{0.58\textwidth}
  \centering
  \includegraphics[width=\textwidth]{%
    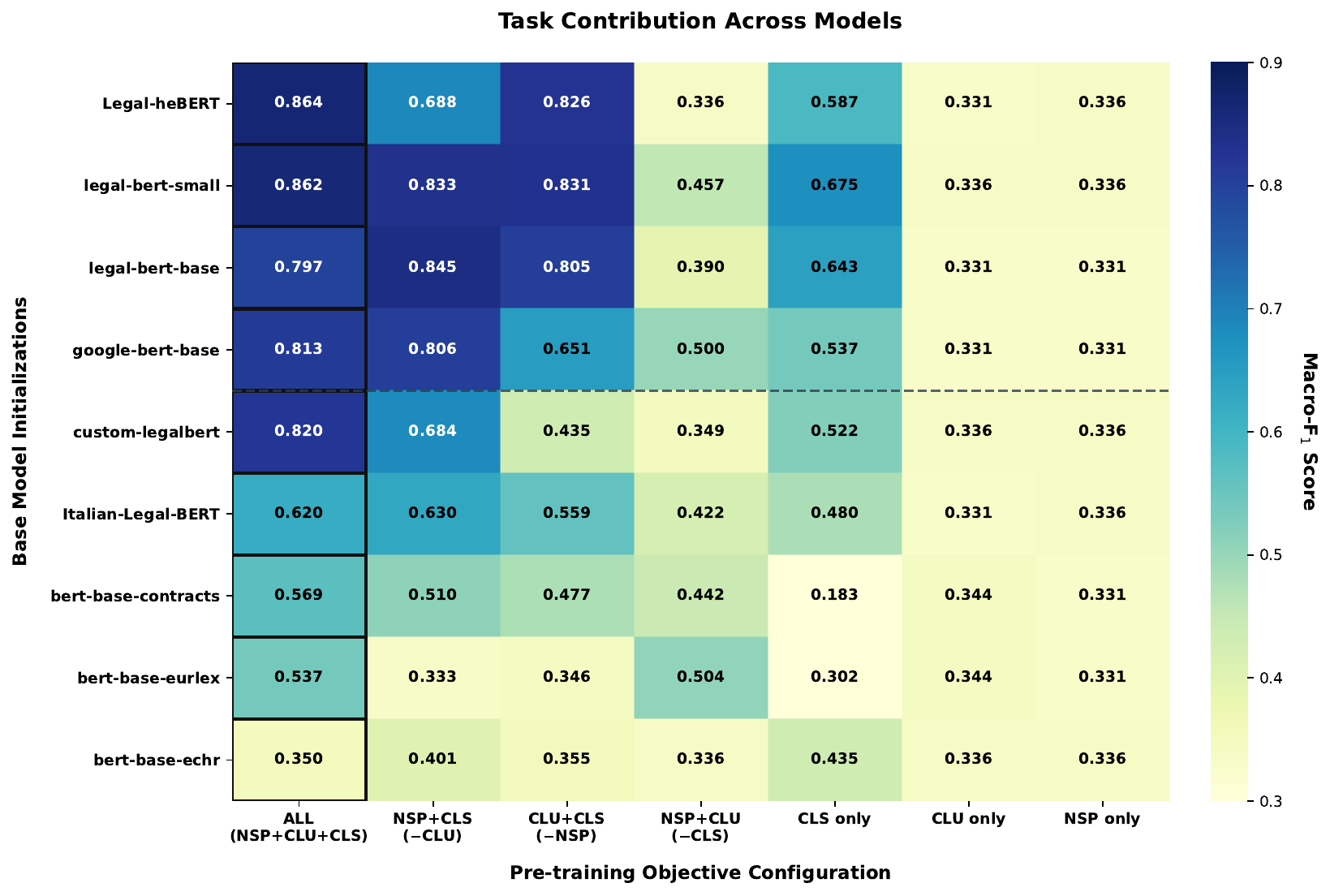}
  \caption{F$_1$-Macro per model across all 7 NSP/CLU/CLS
    configurations. Highlighted ALL column = full multi-task
    objective. Dashed line separates high- and low-performing
    models.}
  \label{fig:ablation_heatmap}
\end{subfigure}
\hfill
\begin{subfigure}[b]{0.38\textwidth}
  \centering
  \includegraphics[width=\textwidth]{%
    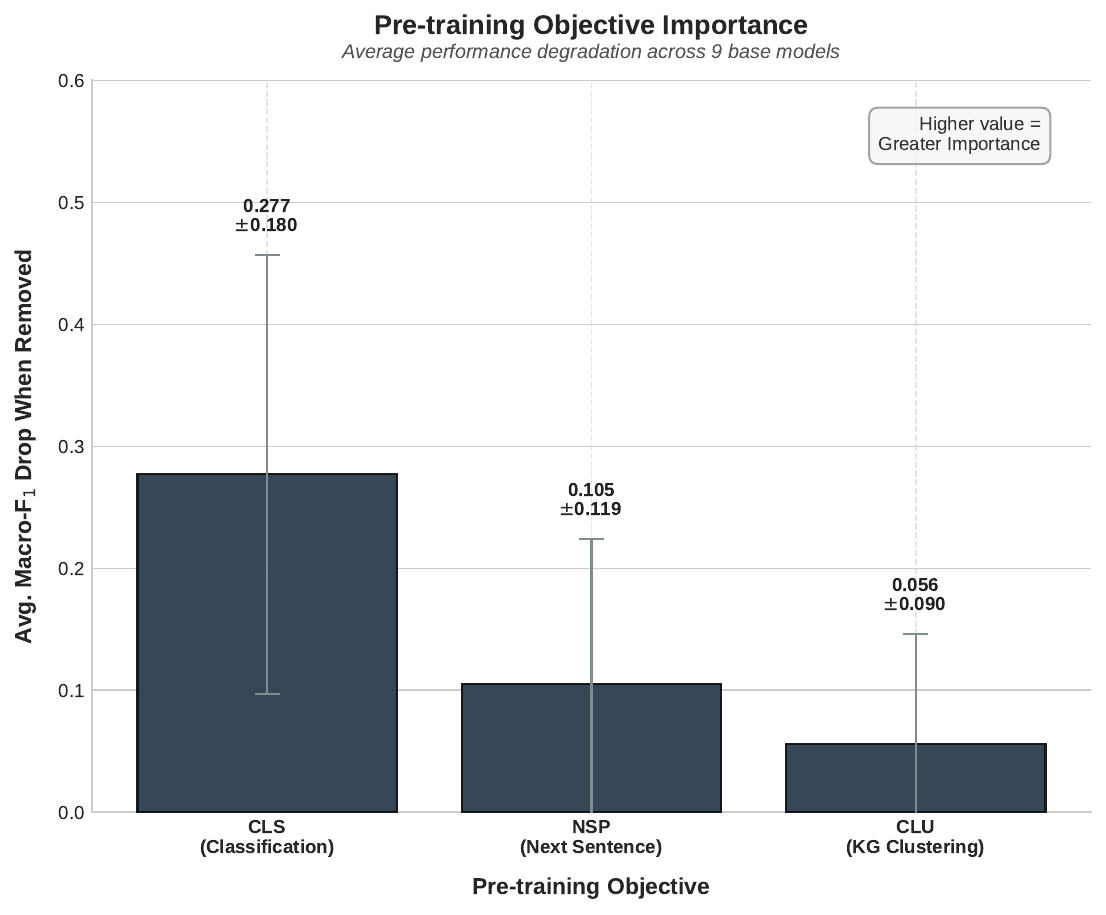}
  \caption{Average $m$-$F_1$ drop when each objective is
    removed across all models. Error bars = std.\ dev.}
  \label{fig:task_importance}
\end{subfigure}
\caption{Ablation study: CLS dominates (avg.\ drop $=0.277$).
\texttt{bert-base-contracts} and \texttt{google-bert-base}
show negative CLU interference, confirming task synergy is
architecture dependent.}
\label{fig:combined_ablation_study}
\end{figure*}

Table~\ref{tab:ablation_full} reports the complete task-contribution ablation
across all models and seven task configurations, supporting the analysis
in Section~\ref{subsec:tasksynergy}. Figure~\ref{fig:combined_ablation_study} visualizes the per-model task
contributions (Figure~\ref{fig:ablation_heatmap}) and the average importance of
each objective (Figure~\ref{fig:task_importance}). Removing CLS causes the largest mean $m\text{-}F_1$ drop ($0.277$), far exceeding the drops from removing NSP ($0.105$) or CLU ($0.056$). Paired
$t$-tests across the models confirm that the two primary objectives
contribute significantly: CLS [$t(8){=}4.35,\,p{<}0.01$] and NSP
[$t(8){=}2.52,\,p{<}0.05$]. The KG-guided CLU objective yields additional gains
for top-performing models such as \texttt{bert-base-echr}
(Table~\ref{tab:stvo_synthetic_data_results}), but its aggregate effect is not
statistically significant [$t(8){=}1.76,\,p{=}0.11$]. This indicates that CLU's
contribution is architecture-dependent and can introduce task interference in
some general-purpose base models (Table~\ref{tab:ablation_full}).

\begin{table*}[t]
\centering
\footnotesize
\setlength{\tabcolsep}{4pt}
\begin{tabular}{l|ccccccc|c}
\toprule
\textbf{Base Model}
  & \textbf{ALL}
  & \textbf{NSP+CLS}
  & \textbf{CLU+CLS}
  & \textbf{NSP+CLU}
  & \textbf{CLS only}
  & \textbf{CLU only}
  & \textbf{NSP only}
  & \textbf{$\Delta$ (w/o CLS)} \\
\midrule
bert-base-echr
  & \textbf{0.864} & 0.688 & 0.826 & 0.336
  & 0.587 & 0.331 & 0.336 & $-$0.528 \\
bert-base-eurlex
  & \textbf{0.862} & 0.833 & 0.831 & 0.457
  & 0.675 & 0.336 & 0.336 & $-$0.405 \\
bert-base-contracts
  & 0.797 & \textbf{0.845} & 0.805 & 0.390
  & 0.643 & 0.331 & 0.331 & $-$0.407 \\
italian-legal-bert
  & \textbf{0.813} & 0.806 & 0.651 & 0.500
  & 0.537 & 0.331 & 0.331 & $-$0.313 \\
custom-legalbert
  & \textbf{0.820} & 0.684 & 0.435 & 0.349
  & 0.522 & 0.336 & 0.336 & $-$0.471 \\
google-bert-base
  & 0.620 & \textbf{0.630} & 0.559 & 0.422
  & 0.480 & 0.331 & 0.336 & $-$0.198 \\
legal-bert-base
  & \textbf{0.569} & 0.510 & 0.477 & 0.442
  & 0.183 & 0.344 & 0.331 & $-$0.127 \\
legal-bert-small
  & \textbf{0.537} & 0.333 & 0.346 & 0.504
  & 0.302 & 0.344 & 0.331 & $-$0.033 \\
Legal-heBERT
  & 0.350 & 0.401 & 0.355 & 0.336
  & \textbf{0.435} & 0.336 & 0.336 & $-$0.014 \\
\midrule
\textbf{Average}
  & \textbf{0.692} & 0.637 & 0.587 & 0.415
  & 0.485 & 0.335 & 0.333 & $-$0.278 \\
\textbf{Std Dev}
  & 0.172 & 0.164 & 0.187 & 0.063
  & 0.147 & 0.005 & 0.003 & 0.183 \\
\bottomrule
\end{tabular}
\caption{Complete ablation: 9 models $\times$ 7 task
configurations. Bold = best per model.
$\Delta$\,(w/o CLS) = ALL $-$ NSP+CLU.}
\label{tab:ablation_full}
\end{table*}


\section{LexGLUE Full Results}
\label{app:lexglue}

Table~\ref{tab:lexglue_direct_comparison_appendix} reports the complete
LexGLUE benchmark results for StVO-pretrained
models against published baselines~\citep{chalkidis-etal-2022-lexglue}
across all seven tasks, covering both $\mu$-$F_1$ and $m$-$F_1$ averaged metrics. Our models improve over the~\citet{chalkidis-etal-2022-lexglue} baselines on four tasks (ECtHR-A,ECtHR-B,UNFAIR-ToS, and
CaseHOLD). See Section \ref{subsec:results_lexglue} for the main analysis.

\begin{table*}[t]
\centering
\resizebox{\textwidth}{!}{%
\begin{tabular}{l|cc|cc|cc|cc|cc|cc|c}
\toprule
\multirow{2}{*}{\textbf{Model}}
  & \multicolumn{2}{c|}{\textbf{ECtHR-A}}
  & \multicolumn{2}{c|}{\textbf{ECtHR-B}}
  & \multicolumn{2}{c|}{\textbf{SCOTUS}}
  & \multicolumn{2}{c|}{\textbf{EUR-LEX}}
  & \multicolumn{2}{c|}{\textbf{LEDGAR}}
  & \multicolumn{2}{c|}{\textbf{UNFAIR-ToS}}
  & \textbf{CaseHOLD} \\
\cmidrule(lr){2-3}\cmidrule(lr){4-5}\cmidrule(lr){6-7}
\cmidrule(lr){8-9}\cmidrule(lr){10-11}\cmidrule(lr){12-13}
\cmidrule(lr){14-14}
  & $\mu\text{-}F_1$ & $m\text{-}F_1$ & $\mu\text{-}F_1$ & $m\text{-}F_1$ & $\mu\text{-}F_1$ & $m\text{-}F_1$
  & $\mu\text{-}F_1$ & $m\text{-}F_1$ & $\mu\text{-}F_1$ & $m\text{-}F_1$ & $\mu\text{-}F_1$ & $m\text{-}F_1$ & $\mu$/m$\text{-}F_1$ \\
\midrule
\multicolumn{14}{c}{\textit{Published Baselines
  \citep{chalkidis-etal-2022-lexglue}}} \\
\midrule
TFIDF-SVM    & 62.6 & 48.9 & 73.0 & 63.8 & 74.0 & 64.4 & 63.4 & 47.9 & 87.0 & 81.4 & 94.7 & 75.0 & 22.4 \\
BERT         & 71.2 & 63.6 & 79.7 & 73.4 & 68.3 & 58.3 & 71.4 & 57.2 & 87.6 & 81.8 & 95.6 & 81.3 & 70.8 \\
RoBERTa      & 69.2 & 59.0 & 77.3 & 68.9 & 71.6 & 62.0 & 71.9 & \textbf{57.9} & 87.9 & 82.3 & 95.2 & 79.2 & 71.4 \\
DeBERTa      & 70.0 & 60.8 & 78.8 & 71.0 & 71.1 & 62.7 & \textbf{72.1} & 57.4 & 88.2 & \textbf{83.1} & 95.5 & 80.3 & 72.6 \\
Longformer   & 69.9 & 64.7 & 79.4 & 71.7 & 72.9 & 64.0 & 71.6 & 57.7 & 88.2 & 83.0 & 95.5 & 80.9 & 71.9 \\
Legal-BERT   & 70.0 & 64.0 & 80.4 & \textbf{74.7} & 76.4 & \textbf{66.5} & 72.1 & 57.4 & 88.2 & 83.0 & 96.0 & 83.0 & 75.3 \\
CaseLaw-BERT & 69.8 & 62.9 & 78.8 & 70.3 & \textbf{76.6} & 65.9 & 70.7 & 56.6 & \textbf{88.3} & 83.0 & 96.0 & 82.3 & 75.4 \\
\midrule
\multicolumn{14}{c}{\textit{Our StVO Multi-Task Models (RQ3)}} \\
\midrule
BERT-EurLex+StVO       & 70.8 & \textbf{65.7}$^\dagger$ & 79.7 & 73.8 & 70.8 & 61.1 & 70.3 & 46.6 & 87.8 & 82.1 & 95.7 & 79.4 & 71.5 \\
BERT-ECHR+StVO         & \textbf{72.7}$^\dagger$ & 61.2 & \textbf{81.0}$^\dagger$ & 73.9 & 71.5 & 62.7 & 69.3 & 44.9 & 87.8 & 81.1 & 95.8 & 82.5 & 71.9 \\
Custom-LegalBERT+StVO  & 71.2 & 61.6 & 78.6 & 70.6 & 68.9 & 48.1 & 67.7 & 43.5 & 87.2 & 81.2 & \textbf{96.8}$^\dagger$ & \textbf{86.0}$^\dagger$ & \textbf{77.0}$^\dagger$ \\
\bottomrule
\end{tabular}}
\caption{Full LexGLUE results across all seven tasks: our
\textit{StVO}-pretrained models vs.\ the published baselines
of~\citet{chalkidis-etal-2022-lexglue}. Scores are $F_1$ (\%), micro ($\mu$) and macro
($m$); CaseHOLD is a single $F_1$ ($\mu{=}m$). \textbf{Bold} marks the best
score per column; $\dagger$ marks an increase over the published baselines. Our
models improve on four tasks (ECtHR-A, ECtHR-B, UNFAIR-ToS, CaseHOLD). The models show negative transfer on SCOTUS, EUR-LEX, and LEDGAR.}
\label{tab:lexglue_direct_comparison_appendix}
\end{table*}

\end{document}